# Geometry-Constrained Monocular Scale Estimation Using Semantic Segmentation for Dynamic Scenes

Hui Zhang, *Member, IEEE,* Zhiyang Wu, *Member, IEEE,* Qianqian Shangguan, Kang An, *Member, IEEE*

***Abstract*—Monocular visual localization plays a pivotal role in advanced driver assistance systems and autonomous driving by estimating a vehicle's ego-motion from a single pinhole camera. Nevertheless, conventional monocular visual odometry encounters challenges in scale estimation due to the absence of depth information during projection. Previous methodologies, whether rooted in physical constraints or deep learning paradigms, contend with issues related to computational complexity and the management of dynamic objects. This study extends our prior research, presenting innovative strategies for ego-motion estimation and the selection of ground points. Striving for a nuanced equilibrium between computational efficiency and precision, we propose a hybrid method that leverages the SegNeXt model for real-time applications, encompassing both ego-motion estimation and ground point selection. Our methodology incorporates dynamic object masks to eliminate unstable features and employs ground plane masks for meticulous triangulation. Furthermore, we exploit Geometry-constraint to delineate road regions for scale recovery. The integration of this approach with the monocular version of ORB-SLAM3 culminates in the accurate estimation of a road model, a pivotal component in our scale recovery process. Rigorous experiments, conducted on the KITTI dataset, systematically compare our method with existing monocular visual odometry algorithms and contemporary scale recovery methodologies. The results undeniably confirm the superior effectiveness of our approach, surpassing state-of-the-art visual odometry algorithms. Our source code is available at https://github.com/bFr0zNq/MVOSegScale.**

***Index Terms*—*geometry-constraint, hybrid method, monocular scale recovery, semantic segmentation, visual odometry.***

## I. INTRODUCTION

VISUAL odometry (VO), or visual localization, is presently a subject of intense exploration as a foundational component of simultaneous localization and mapping (SLAM) in the realms of robotics [1], [2]. Its primary objective is to estimate ego motion from consecutive robot visual signal [3], [4]. In recent years, stereo-based methods [5], [6], [7] have made notable strides due to their proficiency in facile feature matching which is a task that proves challenging in LiDAR-based SLAM approaches [8], [9], [10]. However, stereo methods come with the requirement for additional hardware compared to their monocular counterparts, and their effectiveness heavily relies on baseline calibration.

In practical scenarios, monocular vision systems take precedence due to their cost-effectiveness, simplified mechanical integration, reduced complexity, and resource efficiency. They find extensive applications in various domains such as smartphones, surveillance cameras, robotics, and autonomous vehicles. Consequently, monocular visual odometry (MVO) has garnered more attention compared to multi-sensor approaches.

A limitation of MVO lies in its incapacity to determine the absolute scale of the scene, resulting in what is commonly termed scale ambiguity [11]-[13]. This characteristic induces cumulative errors and inaccurate estimations over time [14]-[16]. To counteract this challenge, additional measurement devices such as Inertial Measurement Units (IMUs) [17], [18], or depth sensors like LiDAR [19], are frequently employed. However, integrating these devices into the estimation system introduces inherent uncertainty, necessitating the adoption of supplementary filters or sophisticated fusion algorithms to effectively manage this uncertainty.

To tackle the issue of scale ambiguity in MVO, two primary approaches are commonly employed: one involves relative scale correction, while the other focuses on absolute scale estimation. Recent advancements in deep learning have shown promising results in addressing the scale ambiguity issue. Liu et al. [20] proposed an unsupervised learning framework fusing visual and inertial measurements for monocular depth estimation, while Wang et al. [21] introduced ScaleNet, an unsupervised scale network for recovering absolute depths. The former initializes the scale for motion estimation in subsequent frames using methods such as perspective-n-points (PnP), maintaining a consistent scale through processes like bundle adjustment (BA), and loop closure (LC) [22], [23]. This method preserves scale when loop closure is detected. However, without loop closure, the algorithm may rely solely on feature tracking, leading to potential scale drift or inconsistencies, impacting the overall accuracy and reliability of the VO system, especially in long-term navigation scenarios.

Popular methods for absolute scale estimation include using references such as camera height [11], [24], [25], baseline distance (in stereo systems), or known object sizes. Among these, the mounted camera height has gained increasing attention as a reliable scale reference during vehicle motion. Recent research has explored learning-based MVO algorithms [26], [27], [28], broadly categorized into two branches. The first treats the MVO problem as an end-to-end process [29], [30], where the pose estimate with absolute scale is directly

This work was supported in part by the National Natural Science Foundation of China (Grant No. 62073245) and the Natural Science Foundation of Shanghai (20ZR1440500), (Hui Zhang and Zhiyang Wu contributed equally to this work.) (Corresponding author: Kang An.)

Hui Zhang, Zhiyang Wu, Qianqian Shangguan and Kang An are with the College of Information, Mechanical and Electrical Engineering, Shanghai Normal University, Shanghai 201418, China.(e-mail: huiz@shnu.edu.cn, 1000512056@smail.shnu.edu.cn, shangguan@shnu.edu.cn, ankang526@foxmail.com).

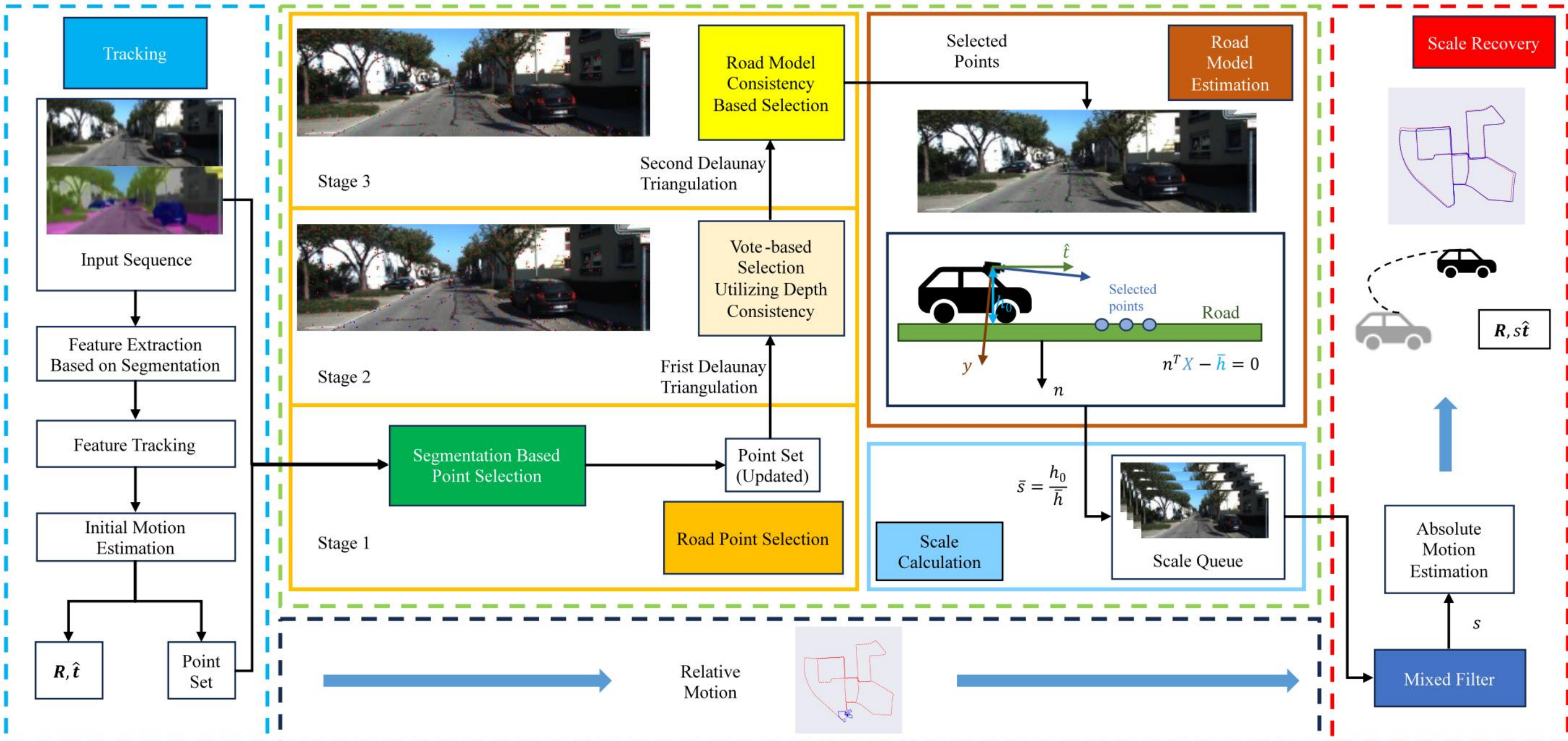

**Fig. 1.** Structure of proposed method. Our approach can be integrated as a separate thread that runs alongside motion estimates. A series of monocular frames serve as the method's input, and dynamic segmentation is used to enhance the feature quality. The initial motion is calculated for each frame using the standard VO approach. The triangulated feature (red point in the Stage 2) is then subjected to a thorough selection process comprising segmentation, depth consistency, and road model consistency to ascertain whether it belongs to the road region. We utilize these features to determine the scale depending on the mounted camera height. The middle result is labeled with blue (Stage 3), and the final selected features are shown in the figure as green dots (Stage 3). To migrate the influence of noise, a mixed filter is proposed for the final scale result to address the absolute motion estimation.

obtained using monocular frames and a corresponding trained model. To address the scarcity of training data, unsupervised methods [25], [31], [32] combine depth prediction and pose estimation. The second branch uses large pretrained depth estimation models to transform the MVO problem into an RGB-D problem, leveraging traditional estimation algorithms [33]. Others predict panoptic-level features to enhance robust matching [34]. However, these learning-based methods often demand computationally intensive resources.

The camera's height serves as a commonly adopted method for scale estimation. This method entails determining the scale by estimating the ratio between the actual camera height ($h_0$) and the calculated one ($\bar{h}$). Accurate estimation of the road model relies on crucial road point selection. Previous studies, as demonstrated in references [11], [12], [35], have utilized homography matrix decomposition to derive the normal vector of the road plane. B. M. Kitt et al. [36] introduce a region of interest (ROI) based method, which confines feature point selection to a fixed frame region. Nonetheless, these methods have limitations in feature point selection, potentially decreasing motion estimation accuracy due to the absence of stable features outside the plane or ROI. In contrast, Lee et al. [37] propose a segmentation-based method that utilizes segmentation in key frames and employs road segmentation labels for road point selection. This method excels even in scenes with moving objects. However, it relies on deep learning-based semantic segmentation, which may exhibit limited generalization performance in real-life scenarios, and its effectiveness might not be guaranteed in all situations. In specific domains, researchers have leveraged environment-specific features to enhance MVO performance. Huang et al. [38] developed FRVO-Mono for railway localization using environment-specific features, and Song et al. [39] presented a visual-inertial fusion method for hyper-redundant robots in constrained environments.

Recent research has also explored advanced techniques for monocular 3D object detection and SLAM in dynamic environments. Yang et al. [40] introduced MonoPSTR, employing a dynamic position and scale-aware transformer for monocular 3-D detection, while Wen et al. [41] proposed CD-SLAM, a real-time stereo visual-inertial SLAM system designed for complex dynamic environments.

To harness the synergies between deep learning and conventional methodologies, effectively overcoming the limitations of prior scale recovery methods and mitigating the impact of dynamic objects on motion estimation, this paper introduces an innovative hybrid approach for scale recovery in MVO, as illustrated in Fig. 1, which can be run in a real time scenario. A lightweight variant of SegNeXt [42], adept at balancing performance and computational complexity for both ego-motion estimation and ground point selection, is employed. The method involves several pivotal steps. Initially, the segmented dynamic object mask is utilized to eliminate unstable features, ensuring a more precise selection of triangulation points for ground plane estimation. Incorporating the ground plane mask generated by SegNeXt, the triangulated points undergo preprocessing. For a more refined feature point selection, Delaunay triangulation [43] partitions planar feature points into triangles. Building upon this, the proposed method validates each triangle to determine whether its constituent feature points belong to the road surface. The road model relies on the validation of road points through the application of random sample consensus (RANSAC) [44], enhancing the robustness of the estimation process by removing outliers. A sliding window method is introduced, enabling online updates of the scale estimate using a mixed filter to migrate noise. In summary, this study makes the following main contributions:

1) We introduce a hybrid scale recovery system designed for the processing of monocular visual signals, strategically leveraging the advantages offered by both deep learning models and geometric constraints. By synergizing these domains, we capitalize on the model's proficiency in eliminating unstable features and incorporate it in the initial screening of ground points. This not only expedites geometric computations but also elevates their precision.
2) We introduce an independent framework capable of seamless integration with existing MVO systems to address the scale ambiguity issue. This framework not only stands alone but also harmonizes with other MVO systems, providing a comprehensive solution to the challenge of scale estimation.
3) We seamlessly incorporate our method into the monocular version of ORB-SLAM3, specifically without loop closure, showcasing its practical applicability in real-world scenarios. The experiments conducted on the KITTI dataset unequivocally highlight its superiority over state-of-the-art algorithms, representing a substantial advancement in the field. This integration not only attests to the versatility of our approach but also establishes it as a leading contender in the domain of monocular visual odometry, surpassing the performance benchmarks set by existing algorithms.

This paper extends our previous investigations [25], [45] by introducing an enhanced feature selection system augmented with semantic segmentation and an improved scale recovery method incorporating geometric constraints. Through the adoption of this hybrid approach, we aim to address the challenges associated with scale recovery in MVO systems. The subsequent sections of this paper follow a structured sequence: Section II provides the research background, Section III details our scale recovery approach, Section IV presents the experimental results using the KITTI dataset, and finally, our conclusions and directions for future work are outlined.

## II. Background

### *A. Semantic Segmentation*

Semantic segmentation stands as a foundational task in computer vision, with convolutional neural networks (CNNs) at the forefront since the introduction of Fully Convolutional Networks (FCN) [46], [47] . However, with the rapid development of Transformers, an increasing number of Transformer-based methods have emerged, surpassing CNN-based approaches in various aspects [48], [49] . Nonetheless, CNNs have proven to be successful in capturing spatial information and achieving high performance with relatively lower computational requirements, which is particularly important in VO systems. Recent methods have also explored integrating attention mechanisms by emulating them through CNNs (e.g., SENet, STN, CBAM), yielding promising results.

SegNeXt stands as the current state-of-the-art segmentation method in the field, employing the widely recognized Encoder-Decoder architecture. In the Encoder section, it innovatively replaces the attention mechanism with an improved multi- scale convolution pattern, akin to LKA. Each module comprises parallel 1x7, 1x11, and 1x21 strip-wise convolutions. Transitioning to the Decoder part, a partially concatenated structure is adeptly utilized to derive the final representation of
features.

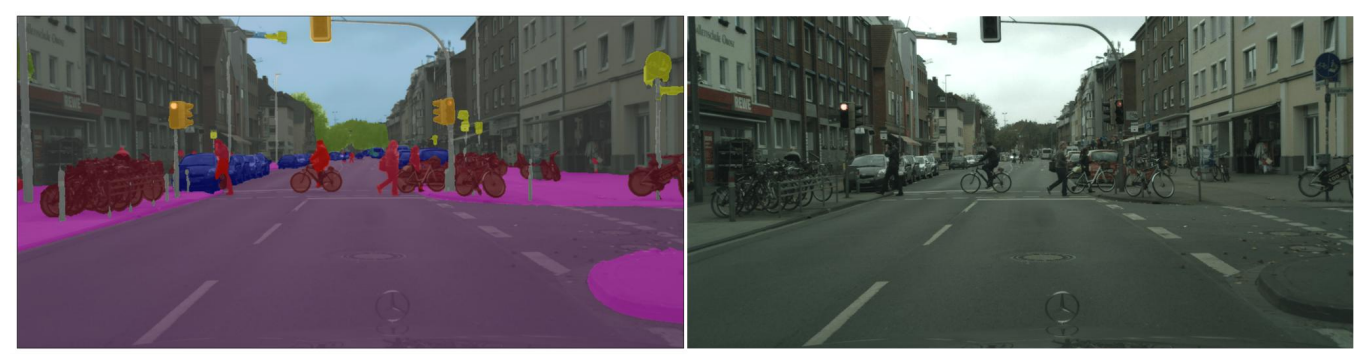

**Fig. 2.** Performance of Selected Segmentation Model (SegNeXt tiny version).

Within the context of this paper, the SegNeXt tiny version is specifically employed for dynamic object and road region prediction. The effectiveness of this chosen model is illustrated in Fig. 2.

### *B. Monocular Motion Estimate*

MVO is geared towards the precise estimation of a camera's trajectory through the analysis of a sequence of images. In this process, MVO seeks to obtain the camera's pose, denoted as $P_t$, from a series of images taken by a single camera. This estimation is achieved by leveraging an initial pose, typically denoted as $P_0$ and extracting information about the camera's motion from consecutive images, such as $I_{t-1}$ and $I_t$. MVO plays a critical role in various applications, including robotics, autonomous navigation, and augmented reality, where it enables precise localization and mapping by continuously tracking the camera's movement based on visual cues. The motion between $I_{t-1}$ and $I_t$ can be defined as:

$$\boldsymbol{T} = \begin{bmatrix} \boldsymbol{R} & \boldsymbol{t} \\ 0 & 1 \end{bmatrix} = \boldsymbol{P}_{t-1}^{-1} * \boldsymbol{P}_t \,, \quad (1)$$

where we have $\boldsymbol{P}$ represents the position of camera, $\boldsymbol{T}$ represents the transformation matrix between two frames, $\boldsymbol{R} \in SO(3)$, which is the rotation matrix describing the camera's rotation from timestamp $t-1$ to $t$, and $\boldsymbol{t} \in R^{3\times1}$ is the translation vector that represent the translation from $t-1$ to $t$. Based on project matrix $\boldsymbol{K}$, we can project the 3D feature points into 2D image points using formula $\boldsymbol{x}_t = \boldsymbol{K}\boldsymbol{X}_t$, where $\boldsymbol{X_t}$ is the notation of selected 3D points and $\boldsymbol{x}_t$ is the corresponding 2D features. Then we can use epipolar constraint for the estimate of $\boldsymbol{T}$:

$$\boldsymbol{x}_{t-1}\boldsymbol{K}^{-T}s\boldsymbol{t}^{\wedge}\boldsymbol{R}\boldsymbol{K}^{-1}\boldsymbol{x}_t = 0, \tag{2}$$

$\wedge$ represent the skew-symmetric operator, which convert the vector into skew-symmetric matrix form, and $s$ is the scale factor. Commonly, we can use the eight-point-algorithm [50] for the decomposition of $\boldsymbol{E} = \boldsymbol{t}^{\wedge}\boldsymbol{R}$.

After estimating the motion, we can recover the 3D coordinate of feature points with a normalized depth using triangulate. In the subsequent frames, 2D-to-3D mapping relationship are provided, enabling the derivation of camera motion $T$ with Perspective-n-Points (PnP) method which optimize the motion estimation by minimizing the reprojection error:

$$\boldsymbol{T} = \underset{T}{\operatorname{argmin}} \sum_i \left|\left|\widehat{\boldsymbol{x}}_{t-1}^i - \boldsymbol{x}_t^i\right|\right|_2 , \tag{3}$$

here $i$ refers to the $i$-th feature point. The project function has following format:

$$\widehat{\boldsymbol{x}}_{t-1}^i = \boldsymbol{K}\left(\boldsymbol{R}\boldsymbol{X}_{t-1}^i + \boldsymbol{t}\right), \tag{4}$$

$||\cdot||_2$ is the $L^2$-norm.

## III. Scale Recovery Approach

The careful selection of feature points is paramount to ensuring precise and robust motion estimation. Various critical aspects must be considered, including the distinctiveness, repeatability, and geometric stability of the chosen feature points. Distinctive feature points exhibit strong gradients or a high corner response, facilitating their reliable detection and tracking. In this section, we present three distinct methodologies for feature selection, each meticulously crafted to support accurate motion estimation and meticulous scale recovery.

Firstly, we introduce the feature refinement method using dynamic segmentation for motion estimation. This approach involves leveraging dynamic segmentation algorithms to refine the detected feature points, enhancing their quality and reliability for motion estimation.

Secondly, we delve into our road point selection algorithms, integral to road model computation. These algorithms focus on selecting the most suitable feature points for road regions, considering their distinctiveness and geometric stability. By concentrating on road points, we elevate the accuracy of road model estimation.

Finally, we employ the RANSAC algorithm to compute the road model, robustly estimating road plane parameters by iteratively fitting the model to the selected road points. To mitigate noise effects, we implement a mixed filter methodology, combining multiple filters. This approach boosts the stability and reliability of both motion and scale estimations.

Through the utilization of these feature selection methods and robust estimation algorithms, our aim is to enhance the accuracy and robustness of motion estimation and scale recovery within the proposed approach.

### A. Feature Refinement by Dynamic Segmentation

As previously mentioned, feature matching is a crucial step in the motion estimation and triangulation processes. In the realm of feature point extraction, Various methods have been employed for feature point extraction [51], [52], [53], [54] . Traditional methods like SIFT [55], [56] , SURF [57] and BRISK [58] have long been used for their ability to detect scale-invariant and robust key points. The introduction of ORB [24] further enhanced efficiency and rotational invariance.

Deep learning has introduced novel methods such as SuperPoint [59], which exhibits impressive performance and rapid feature detection. D2-Net is notable for its dense and high-quality feature point matching capabilities, while LF-Net excels in combining local and global context for challenging scenarios. Despite their strengths, these deep learning-based methods come with computational resource requirements during training and deployment and heavily rely on labeled data. They can also be sensitive to variations in the training data distribution, posing challenges in diverse real-world conditions. Furthermore, existing algorithms, including the mentioned ones, struggle with dynamic objects. When key points are on moving objects, the performance of motion estimation and related tasks can significantly degrade, impacting accuracy.

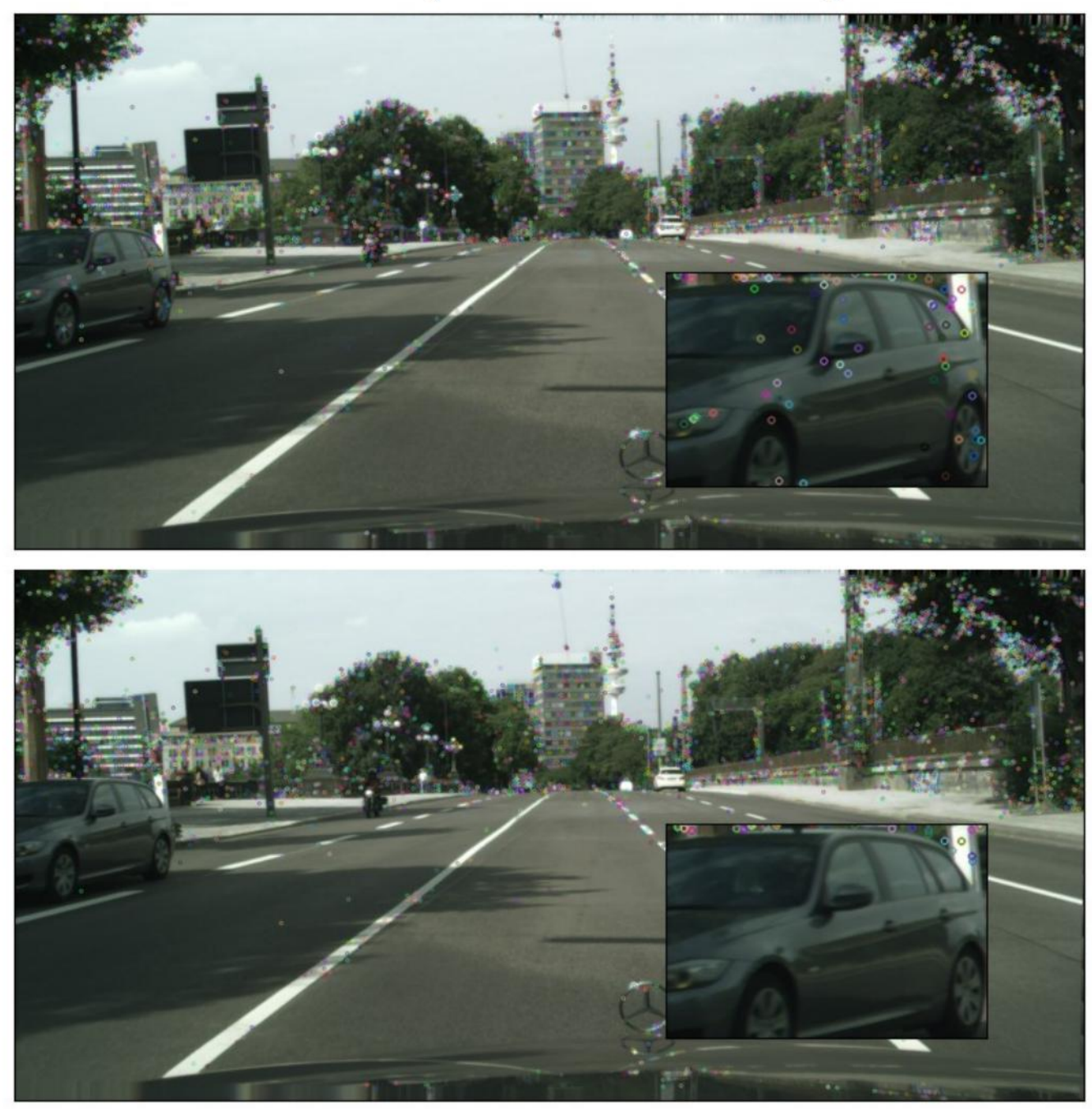

**Fig. 3.** Feature refinement using dynamic segmentation. Features localize on dynamic object is decreased significantly.

To tackle these challenges, we employ a dynamic segmentation method to effectively reduce the influence of dynamic objects. By integrating the mask generated through this approach with the SIFT detector, we notably decrease the number of features on dynamic objects. This strategic combination helps address challenges posed by dynamic objects and contributes to improved accuracy in feature matching, as visually depicted in Fig. 3.

### *B. Road Surface Model Estimation*

The estimation of the road surface model involves a two-stage process designed to effectively identify and select 3D points belonging to the road surface. In the initial stage, we employ Delaunay triangulation to partition the features along with their corresponding 3D points into multiple triangular regions. This integrative approach incorporates both depth constraints and road segmentation data obtained from SegNeXt. By synergizing depth consistency and road segmentation results, we can precisely pinpoint the road points.

Following the initial selection, a subsequent refinement of road point selection occurs by applying Delaunay triangulation to the remaining points. This refinement step enhances the accuracy of triangulation based on the revised set of points. Additionally, we remain committed to excluding road point outliers by assessing their consistency with the established road model.

To provide further clarity, we introduce two meticulously designed algorithms aimed at enhancing the precision and robustness of our road point selection process. The crux of Algorithm 1 lies in the careful consideration of both depth constraint and segmentation information. Algorithm 2 is a geometrical consistency-driven road point refinement method. These algorithms work together, providing a comprehensive and robust approach to road point selection.

1) **Depth-Consistency-Based Road Point Selection with Segmentation Preprocess**

As described in section III-A, after initial motion is estimated, we can obtain triangulated 3D points for corresponding 2D features that track successfully between two frames. Take the remaining features set $x_{t-1}, X_{t-1}$, we can impose the following constraints.

First, under the KITTI camera coordination, we can simply denote the $y$ - coordinate of the 3D point as $\overline{h}$ which is a calculated height of points and we denote the $z$ – coordinate as $\overline{d}$ as it is the depth of the 3D points, further we denote the pixel coordinate of the 2D features as $(u,v)$, if the feature points belong to the road, then it has following constraint:

$$v = \frac{P_y \cdot f_y}{P_z} + c_y \leftrightarrow \overline{d} = \frac{\overline{h} \cdot f_y}{v - c_y}, \quad (5)$$

where the former part of the formula is the project function in $y$ – coordinate, $f_y, c_y \in K$ and $\boldsymbol{X^i} \in \boldsymbol{X_{t-1}}$. Former formula can be concluded as $\overline{d} \propto \frac{1}{v}$. Moreover, when considering any pair of ground plane feature points $X^i, X^j$ with corresponding pixel coordinates $(v_i, v_j)$ the following constraint holds:

$$\sigma(i,j) = \left(v_i\text{-}v_j\right)\left(\overline{d}_i\text{-}\overline{d}_j\right) \le 0, \quad (6)$$

if $\sigma > 0$, it signifies that at least one of the features is not associated with the road surface. However, the aforementioned equation in isolation cannot ascertain which feature should be excluded. To tackle this challenge, we introduce selection mechanisms grounded in voting and segmentation, as elaborated in Algorithm 1.

2) **Geometrical Consistency-Driven Road Point Refinement**

Continuing the selection process, we identify the triangles that adhere to the road model consistency. Using the results generated by Algorithm 1, we perform Delaunay triangulation on the remaining points, resulting in a set of triangles denoted as $\boldsymbol{\Delta}$. Subsequently, we engage in the derivation of the geometric representation for each triangle area, denoted as $\Delta_i$ by solving the equation:

$$\boldsymbol{n_i^{\mathrm{T}}} \cdot \boldsymbol{X^i} - \overline{h}_i = 0, \quad (7)$$

**Algorithm 1:** Feature Choice Grounded in Depth Consistency and Road Segmentation.

**Input:** Road mask $Mask_{t-1}^R$, Set of 3D points $\boldsymbol{X} = \{X^1, X^2, \ldots, X^n\}$ with corresponding pixel coordinate $(u_i, v_i)$.
**Output:** Selected points set $\boldsymbol{\Omega}$.

**for** $X^i \in \boldsymbol{X}$ **do**
  **if** $Mask_{t-1}^R[u_i, v_i] \neq Feature_{road}$ **then**
    delete $X^i$ from $\boldsymbol{X}$
  **end**
**end**
Establish a criterion $\boldsymbol{\beta_a}$ for assessing whether points in set are part of ground plane.
Get triangles set $\boldsymbol{\Delta} = \{\Delta_1, \Delta_2, \ldots, \Delta_n\}, \Delta_i = \{X^i, X^j, X^k\}$ using Delaunay Triangulate.
**for** $\Delta_i \in \boldsymbol{\Delta}$ **do**
  $vote\{\Delta_i\} = 0$
  **for** $\{X^i, X^j\} \in \Delta_i$ **do**
    Calculate $\sigma$ using (6)
    **if** $\sigma \le 0$ **then**
      $vote\{X^i, X^j\} = vote\{X^i, X^j\} + 1$
    **else**
      $vote\{X^i, X^j\} = vote\{X^i, X^j\} - 1$
    **end**
  **end**
**end**
**for** $X^i \in \boldsymbol{X}$ **do**
  **if** $vote\{X^i\} \ge \boldsymbol{\beta_a}$ **then**
    $\boldsymbol{\Omega} \leftarrow X^i$
  **end**
**end**
**return** $\boldsymbol{\Omega}$

where $X^i \in \Delta_i$, we have four variables and only three constraints in this equation. To obtain a unique solution, we introduce two additional constraints ( $||\boldsymbol{n_i}|| = 1\, and\, n_{iy} > 0$ ). After obtaining the geometric descriptions of each triangular region, we estimate pitch angle of each triangular region with following formula:

$$\theta_i = arcsin\left(\frac{n_{i2}}{|\boldsymbol{n_i}|}\right), \tag{8}$$

here, $n_{i2}$ corresponds the second component of $\boldsymbol{n_i}$. Next, we derive the approximate road plane pitch angle $\theta_r$ from the motion $\hat{t}$ the mounted camera's pitch angle $\theta_{Cam}$:

$$\theta_{\hat{t}} = \begin{cases} arcsin\left(-\frac{\hat{\boldsymbol{t}}_2}{|\hat{\boldsymbol{t}}|}\right) & if \quad |\hat{t}| \neq 0 \\ NaN & if \quad |\hat{t}| = 0 \end{cases}, \tag{9}$$

$$|\theta_{Cam}| = \begin{cases} \left|arctan\left(-\frac{R_{32}}{R_{33}}\right)\right| & if \quad R_{33} \neq 0 \\ \frac{\pi}{2} & if \quad R_{33} = 0 \end{cases}, \tag{10}$$

**Algorithm 2:** Feature Selection Based on Road Model.

**Input:** 3D points set $\boldsymbol{X} = \{X^1, X^2, \ldots, X^n\}$with corresponding pixel coordinate $(u_i, v_i)$, the road model $n_i^T \cdot X^i - \bar{h_i} = 0$, the motion vector $\hat{t}$.

**Output:** Selected points set $\boldsymbol{\Omega}$.

```
Establish a criterion β_b for assessing whether points in
  set are part of ground plane.
Get triangles set
  Δ = {Δ_1, Δ_2, ..., Δ_n}, Δ_i = {X^i, X^j, X^k} using
  Delaunay Triangulate.
Calculate θ_t̂ using (8).
Calculate θ_r using θ_r = θ_t̂ − π/2.
for Δ_i ∈ Δ do
    Obtain the norm n_i and height h̄_i from (7)
    Calculate θ_i using (8)
    if ||θ_r − θ_i|| < β_b then
        for X^i ∈ Δ_i do
            if X^i ∉ Ω then
                Ω ← X^i
            end
        end
    end
end
return Ω
```

we simplify the estimation of the road plane pitch angle by setting $\theta_{Cam} = 0$. As a result, the estimated pitch angle of the road normal, $\theta_r$, is calculated as $\theta_r = \theta_{\hat{\boldsymbol{t}}} - \frac{\pi}{2}$ because when the vehicle is running along the road, the motion vector $\hat{t}$ is tangential to the ground plane. All triangular areas are constrained by the road model using the following formula:

$$\left||\theta_r - \theta_i|\right| < \theta_0, \tag{11}$$

$\theta_0$ represents the angle criterion, a pivotal parameter in our methodology, which we empirically set to 5 degrees during our experimental trials. Only triangular regions that adhere to this constraint will be retained. The specific calculations are detailed in Algorithm 2.

### *C. Scale Calculation*

By section III-B, we exclusively utilize the feature points that satisfy the criteria outlined in Algorithm 1 and Algorithm 2 for the computation of road models and scales. Our principal methodology centers around a plane fitting algorithm. Utilizing this method, we can calculate the camera height within a normalized scale. To ensure the accuracy and robustness of this scale estimation process, we incorporate a floating window mechanism to preserve the calculated scale values, subsequently applying a mixed filter. This comprehensive approach enhances the precision of the scale estimation.

We assume the road geometrical model in frame $I_t$ can be written as:

$$\boldsymbol{n_t^T X_t} - \bar{\mathrm{h}}_{\mathrm{t}} = 0, \tag{12}$$

where $\boldsymbol{n_t}$ is the road plane normal, and $h_t$ is the calculated height in a relative scale. The scale can be recovered with:

$$s_t = \frac{h_0}{\bar{h}_t}, \tag{13}$$

here $h_0$ is the measured camera height. To ensure the accuracy of our road plane estimation, we employ the RANSAC method, a robust statistical technique that relies on verified road points. Nevertheless, as a precautionary measure, if the count of selected feature points drops below a predefined threshold, we choose to bypass the RANSAC step. In these instances, the road model maintains its consistency with the one derived from the preceding validated frame. For our experimental configuration, this threshold has been empirically established at 12.

To further reduce the influence of scale noise, we maintain a queue of post $q$ scales and apply a gaussian filter on it with $\sigma = 5$, and then apply a median filter when recover current scale. We make an assumption that the vehicle speed does not exhibit rapid changes within a short time window. The calculation procedure is outlined below:

$$\boldsymbol{\Omega_s} = \{s_{t-q+1}, s_{t-q+2}, \ldots, s_t\}, \tag{14}$$

$$s_i = \frac{1}{\sqrt{2\pi\sigma^2}} exp\left(-\frac{s^2}{2\sigma^2}\right), s_i \in \boldsymbol{\Omega_s}, \tag{15}$$

$$s_t = median(\boldsymbol{\Omega_s}), \tag{16}$$

we set the window size of 5 in our experiments.

## IV. Experiment

To evaluate the performance of our novel approach, we conduct a comprehensive set of experiments using the MVO system based on ORB-SLAM3. Our proposed method operates in parallel with the frame-to-frame motion estimation in VO as an independent thread within the system architecture. The assessment utilizes the extensively adopted KITTI dataset [60], [61] , comprising 22 sequences spanning diverse urban, rural, and highway environments. For a quantitative evaluation of scale recovery performance, we employ two primary metrics known as relative pose error (RPE) [60] and absolute trajectory error (ATE) [62].

Our experimental evaluation focuses on four main aspects. Firstly, we integrate our method with existing VO systems to conduct both quantitative and qualitative assessments of the achieved performance enhancements. Secondly, we compare our method against recently proposed monocular scale recovery methods. Thirdly, we utilize monocular version of ORB-SLAM3 (without loop closure) as the initial ego-motion estimation method and compare our approach with state-of-the-art VO algorithms. Lastly, we conduct an ablation experiment, aiming to dissect and scrutinize the individual impact and effectiveness of each module within our proposed approach.

TABLE I
COMPARISON BETWEEN OUR SCALE RECOVERY (SR) METHOD AND LC ON ORB-SLAM3

| Seq | Runnning distance(m) | ORB+LC ATE(m) | ORBnoLC+Our SR ATE(m) |
|---|---|---|---|
| 00 | 3724 | 7.59 | **6.44** |
| 02 | 5067 | 19.21 | **6.54** |
| 03 | 561 | **0.94** | 2.65 |
| 04 | 394 | **1.03** | 1.19 |
| 05 | 2206 | 5.64 | **3.18** |
| 06 | 1233 | 12.36 | **4.39** |
| 07 | 695 | **2.02** | 3.65 |
| 08 | 3223 | 89.91 | **4.33** |
| 09 | 1705 | 6.12 | **4.41** |
| 10 | 920 | 4.91 | **2.70** |
| Mean | 1972.80 | 14.97 | **3.94** |

TABLE II
COMPARISON BETWEEN OUR SCALE RECOVERY (SR) METHOD AND ROI-BASED METHOD ON MONOVO

| Seq | Rotation error RPE(deg/m) | MonoVO RPE(%) | MonoVO +ROI RPE(%) | MonoVO +Our SR RPE(%) |
|---|---|---|---|---|
| 00 | 0.0040 | 13.29 | 3.86 | **1.67** |
| 02 | 0.0037 | 11.28 | 3.26 | **1.81** |
| 03 | 0.0024 | 7.07 | 3.50 | **1.20** |
| 04 | 0.0023 | 9.20 | 4.22 | **0.90** |
| 05 | 0.0042 | 9.86 | 5.24 | **1.60** |
| 06 | 0.0035 | 2.99 | 5.76 | **1.88** |
| 08 | 0.0040 | 11.37 | 4.14 | **1.67** |
| 09 | 0.0026 | 10.83 | 3.39 | **1.78** |
| 10 | 0.0036 | 16.78 | 4.96 | **1.99** |
| Mean | 0.0034 | 10.30 | 4.26 | **1.67** |

Through these comprehensive experiments, our goal is to demonstrate the effectiveness and superiority of our scale recovery method in augmenting the performance of VO systems.

*A. Enhancements over Open-Source VO Systems*

To assess the efficacy of our method, we combine it with well-established feature-based localization algorithms, specifically MonoVO and ORB-SLAM3. Subsequently, we conduct a rigorous evaluation, comprising both quantitative and qualitative analyses, to gauge the extent of performance improvements achieved through our introduced scale recovery method.

1) **Enhancements in ORB-SLAM3**

In the initial phase, our seamlessly integrated scale recovery method becomes an integral component of the ORB-SLAM3 framework (Fig. 4). This figure vividly illustrates the utilization of the ORB-SLAM3 trajectory, initially obtained without loop closure, as the foundational estimate. Subsequently, our scale recovery approach is systematically employed following a meticulous filtering process.

To gauge performance, we conduct a quantitative comparison with the ORB-SLAM3 monocular loop closure trajectory. Given the inherent absence of scale recovery in the original ORB-SLAM3 monocular implementation, an initial alignment process is undertaken. This involves the careful adjustment of the ORB-SLAM3 monocular loop closure trajectory through a comprehensive 7-degree-of-freedom (7-DOF) transformation, covering translation, rotation, and scale parameters.

The results of our exhaustive comparative analysis, meticulously outlined in Table I, overwhelmingly indicate the superior performance of our scale recovery method compared to the ORB-SLAM3 monocular loop closure trajectory. Across the majority of instances, our approach not only demonstrates enhanced localization accuracy and efficiency but also displays the capability to maintain performance within acceptable margins. It's crucial to note that in a small subset of cases, our method may exhibit slightly inferior performance when juxtaposed with the ORB-SLAM3 trajectory. Nevertheless, these minor deviations remain well within acceptable margins and do not compromise the overall efficacy and robustness of our approach.

2) **Enhancements in MonoVO**

MonoVO, an open-source MVO system built upon OpenCV, leverages the FAST detector for feature extraction and tracks features using Lucas-Kanade optical flow. While MonoVO demonstrates proficiency in estimating rotation and translation through the essential matrix with RANSAC, it encounters limitations in scale calculation, a pivotal aspect of MVO. To overcome this shortfall, our proposed method has been seamlessly integrated with MonoVO. Notably, as elucidated in [25], we employ a feature point scattering technique to augment the overall system performance significantly. The results, systematically presented in Table II, unequivocally reveal a substantial improvement over the original MonoVO. For instance, in sequence 00, MonoVO records a RPE of 13.29%, whereas MonoVO-ROI, incorporating our feature point scattering technique, achieves a markedly reduced RPE of 3.86%. Importantly, with the application of our scale recovery method, the RPE further diminishes to a remarkable 1.67%, signifying a substantial advancement in overall performance.

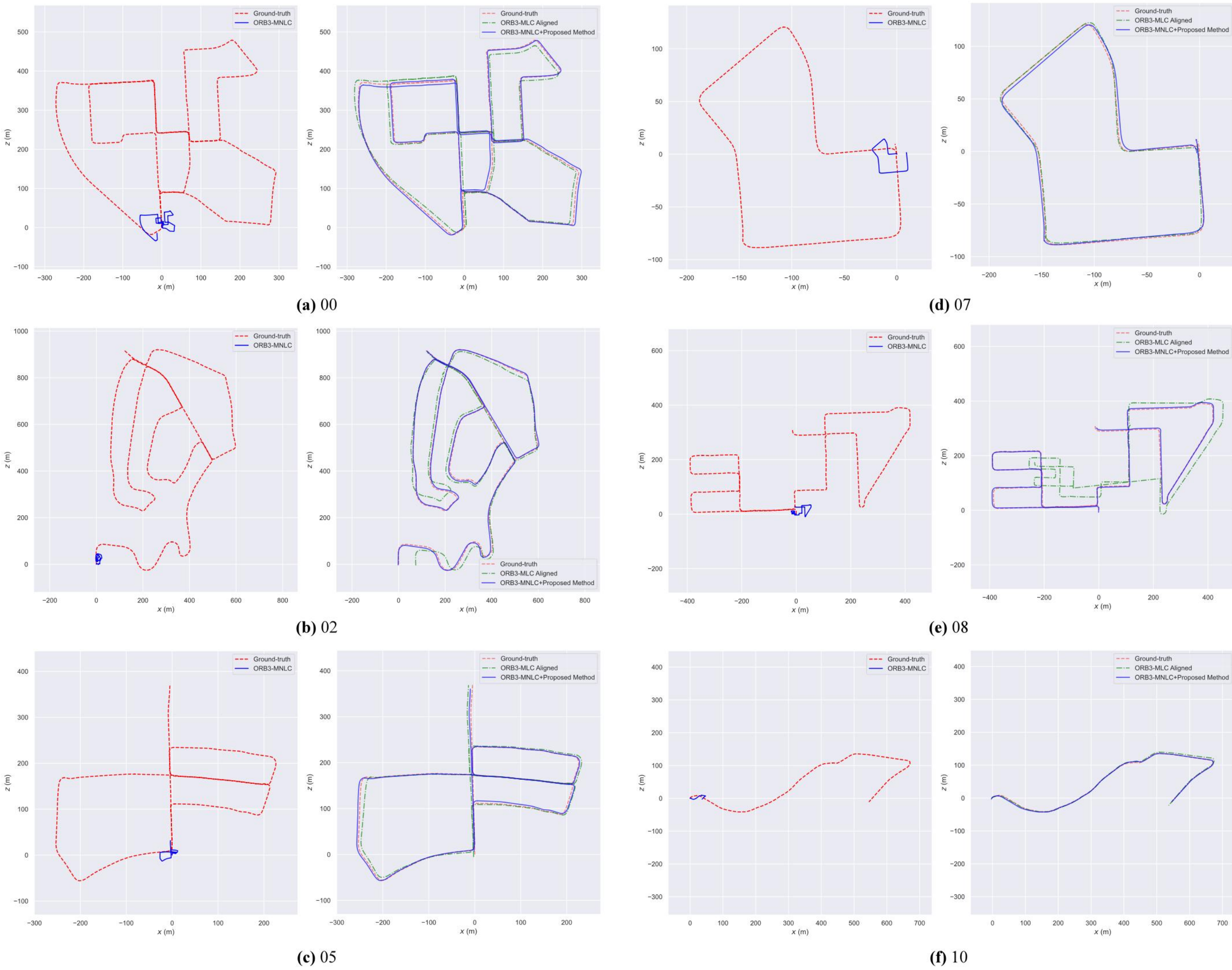

**Fig. 4.** Improvement on ORB-SLAM3 (without loop closure). KITTI sequences 00, 02, 05, 07, 08, 10 are shown in figure, first column of each subplot is the initial motion estimation of ORB-SLAM3 monocular version and the second column is motion after scale recovery. The trajectory estimated by ORB-SLAM3 M (monocular) LC (loop closure) is 7-DOF aligned.

TABLE III
COMPARISON WITH OTHER STATE-OF-THE-ART VO ALGORITHMS

| Seq | Zhou et al. [24] | | Campos et al. [20] | | Lee et al. [30] | | Tian et al. [21] | | Zhan et al. [3] | | Geiger et al. [5] | | Proposed Method | |
|---|---|---|---|---|---|---|---|---|---|---|---|---|---|---|
| | Trans % | Rot (deg/m) | Trans % | Rot (deg/m) | Trans % | Rot (deg/m) | Trans % | Rot (deg/m) | Trans % | Rot (deg/m) | Trans % | Rot (deg/m) | Trans % | Rot (deg/m) |
| 00 | 2.17 | 0.0039 | 3.18 | 0.0040 | 4.42 | 0.0150 | **1.41** | 0.0054 | 2.25 | 0.0058 | 2.32 | 0.0109 | 1.66 | 0.0026 |
| 01 | - | - | - | - | 6.91 | 1.0140 | - | - | - | - | - | - | - | - |
| 02 | - | - | 4.88 | 0.0032 | 4.77 | 0.0168 | 2.18 | 0.0046 | 3.60 | 0.0052 | 2.01 | 0.0074 | **1.38** | 0.0025 |
| 03 | 2.7 | 0.0044 | **0.90** | 0.0013 | 8.49 | 0.0192 | 1.79 | 0.0041 | 2.67 | 0.0050 | 2.32 | 0.0107 | 1.93 | 0.0013 |
| 04 | - | - | **1.07** | 0.0015 | 6.21 | 0.0069 | 1.91 | 0.0021 | 1.43 | 0.0029 | 0.99 | 0.0081 | 1.22 | 0.0009 |
| 05 | - | - | 2.51 | 0.0040 | 5.44 | 0.0248 | 1.61 | 0.0064 | 1.15 | 0.0030 | 1.78 | 0.0098 | **1.12** | 0.0022 |
| 06 | - | - | 5.75 | 0.0015 | 6.51 | 0.0222 | 2.03 | 0.0044 | **1.03** | 0.0026 | 1.17 | 0.0072 | 1.69 | 0.0024 |
| 07 | - | - | **1.56** | 0.0037 | 6.23 | 0.0292 | 1.77 | 0.0230 | **0.93** | 0.0029 | - | - | 1.81 | 0.0031 |
| 08 | - | - | 22.97 | 0.0077 | 8.23 | 0.0243 | 1.51 | 0.0076 | 2.23 | 0.0030 | 2.35 | 0.0104 | **1.42** | 0.0028 |
| 09 | - | - | 2.41 | 0.0027 | 9.08 | 0.0286 | 1.77 | 0.0118 | 2.47 | 0.0030 | 2.36 | 0.0094 | **1.64** | 0.0025 |
| 10 | 2.09 | 0.0054 | 2.02 | 0.0026 | 9.11 | 0.0322 | **1.25** | 0.0031 | 1.96 | 0.0031 | 1.37 | 0.0086 | 1.48 | 0.0024 |
| Mean | 2.32 | 0.045 | 4.72 | 0.0032 | 6.86 | 0.0212 | 1.72 | 0.0068 | 1.97 | 0.0036 | 2.02 | 0.0095 | **1.54** | 0.0023 |

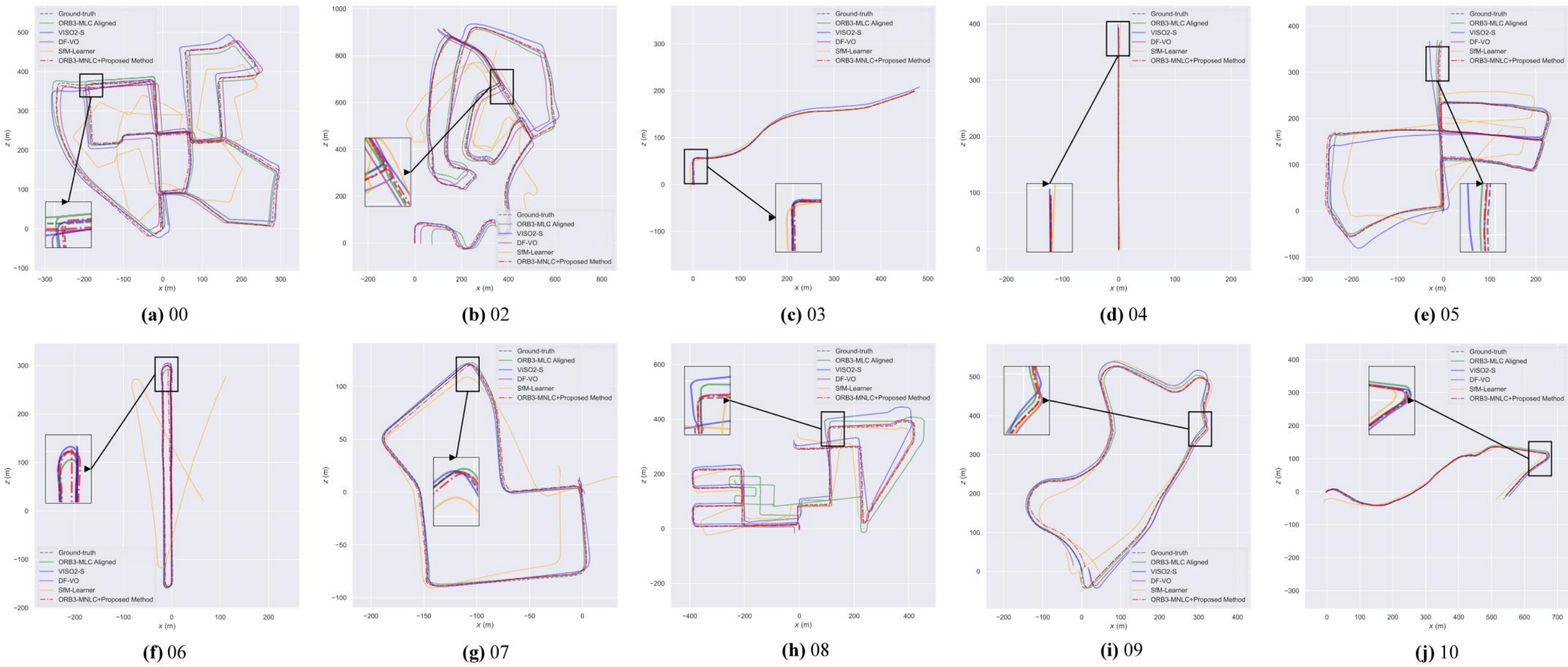

(a) 00 (b) 02 (c) 03 (d) 04 (e) 05

(f) 06 (g) 07 (h) 08 (i) 09 (j) 10

**Fig. 5.** Comparison with other state-of-the-art VO methods. We utilize ORB-SLAM3 monocular (without loop closure) as our initial motion. The gray dashed line in each subfigure is the ground truth, and the red dash-dotted line is the trajectory with our scale recovery method. VISO2-S represent the stereo version of LibVISO2 and the ORB-SLAM3 M (monocular) LC (loop closure) is aligned with 7-DOF alignment algorithm.

### *B. Comparison with State-of-the-Art Algorithms*

In this section, we systematically assess the efficacy of our proposed methodology through its integration with ORB-SLAM3 (without loop closure) and subsequent comparison with contemporary scale recovery techniques [25], [35] and state-of-the-art VO algorithms [3], [5], [25], [24], [24], [30].

Firstly, the evaluation is conducted using the ATE metric.

TABLE IV

COMPARISON WITH OTHER SCALE RECOVERY METHOD USING ORB-SLAM3 MONOCULAR (WITHOUT LOOP CLOSURE) INITIAL MOTION ESTIMATION

| Seq | J. Lee et al. [35] ATE(m) | H. Zhang et al. [24] ATE(m) | ORBnoLC+Our SR ATE(m) |
|---|---|---|---|
| 00 | 6.79 | 8.45 | **6.44** |
| 02 | 13.05 | 11.74 | **6.54** |
| 03 | **1.30** | 1.22 | 2.65 |
| 04 | 1.27 | 1.74 | **1.19** |
| 05 | **2.88** | 4.21 | 3.18 |
| 06 | - | 11.17 | **4.39** |
| 07 | **2.02** | 5.51 | 3.65 |
| 08 | 12.97 | 5.74 | **4.33** |
| 09 | - | 4.91 | **4.41** |
| 10 | - | 4.48 | **2.70** |
| Mean | 5.75 | 5.91 | **3.94** |

The outcomes presented in Table IV demonstrate the superiority of our approach over recent scale recovery techniques, exhibiting relatively low average ATE values. Notably, our method demonstrates superior performance in long sequences of the KITTI dataset, such as sequence 00 (3.7 km), 02 (5.1 km), and 08 (2.8 km), indicating its robust scale recovery process.

Additionally, we conduct a comparative analysis of our approach against state-of-the-art VO methodologies, employing the RPE metric on sequences 00 and 02-10. As illustrated in Table III, our method consistently maintains a lowest average error in comparison to various VO systems, including stereo methods.

These meticulous evaluations collectively underscore the effectiveness of our scale recovery method, not only showcasing enhanced performance in MVO but also positioning it in league with sophisticated stereo-based SLAM algorithms.

### *C. Ablation and Performance Experiment*

In this section, we delve into a comprehensive analysis of each module constituting our proposed method, segmented into three distinct stages: feature extraction, feature selection, and filtering. The initial motion estimation is facilitated by MonoVO, incorporating scattered features. To ensure the consistency and comparability of our analysis, experiments are conducted on sequences 00 and 02-10.

Primarily, we subject the road model consistency algorithm and the vote-based depth consistency algorithms to a quantitative test in an incremental manner. The results, thoughtfully presented in Table V, serve as a testament to the effectiveness of our feature selection strategies. The average RPE value of MonoVO significantly decreases from 10.30% to 2.52% and 2.33% with the incorporation of the Road Model consistency (RM) and Vote-based Feature Selection (VS) modules, respectively. These improvements across almost all sequences underscore the reliability and impact of these modules, derived from robust methodologies.

Subsequently, we turn our attention to the evaluation of different detectors, recognizing the pivotal role of feature extraction in VO. As highlighted in Table VI, the results accentuate the advantages of the SIFT and SuperPoint algorithms. The SIFT corner point extraction algorithm, in particular, achieves the lowest RPE value, showcasing its effectiveness in robust feature detection. Consequently, in this paper, we adopt the SIFT corner point extraction algorithm to enhance feature detection.

TABLE V
ABLATION EXPERIMENT OF PROPOSED ROAD MODEL CONSISTENCY (RM) AND VOTE-BASED FEATURE SELECTION UTILIZE DEPTH CONSISTENCY (VS)

| Seq | Rotation error (deg/m) | No selection % | RM % | RM+VS % |
|---|---|---|---|---|
| 00 | 0.0040 | 13.29 | 2.13 | **2.11** |
| 02 | 0.0037 | 11.28 | 2.17 | **1.97** |
| 03 | 0.0032 | 7.07 | 1.78 | **1.74** |
| 04 | 0.0019 | 9.20 | **1.23** | 1.41 |
| 05 | 0.0039 | 9.86 | 1.91 | **1.66** |
| 06 | 0.0027 | 2.99 | 2.44 | **1.99** |
| 07 | 0.0274 | 12.30 | 6.23 | **5.86** |
| 08 | 0.0040 | 11.37 | 2.41 | **2.28** |
| 09 | 0.0027 | 10.83 | 2.08 | **1.73** |
| 10 | 0.0041 | 16.78 | 2.74 | **2.51** |
| Mean | 0.0058 | 10.30 | 2.51 | **2.33** |

TABLE VI
ABLATION EXPERIMENT OF DIFFERENT FEATURE DETECTOR

| Seq | Rotation error (deg/m) | BRISK % | AKAZE % | Super point % | SIFT % |
|---|---|---|---|---|---|
| 00 | 0.0040 | 2.49 | 2.25 | **1.78** | 2.11 |
| 02 | 0.0037 | 2.09 | 2.06 | 2.05 | **1.97** |
| 03 | 0.0032 | 1.64 | **1.36** | 1.99 | 1.74 |
| 04 | 0.0019 | 3.66 | 2.05 | 1.61 | **1.41** |
| 05 | 0.0039 | 1.81 | 2.05 | 2.76 | **1.66** |
| 06 | 0.0027 | 3.34 | 2.52 | **1.49** | 1.99 |
| 07 | 0.0274 | 8.77 | 6.20 | **5.59** | 5.86 |
| 08 | 0.0040 | 2.80 | **2.25** | 2.37 | 2.28 |
| 09 | 0.0027 | 1.52 | 2.24 | 2.26 | 1.73 |
| 10 | 0.0041 | 1.85 | 2.42 | 2.17 | 2.51 |
| Mean | 0.0058 | 3.00 | 2.54 | 2.41 | 2.33 |

TABLE VII
ABLATION EXPERIMENT OF PROPOSED SEGMENTATION COMPONENT

| Seq | Rotation error (deg/m) | W/O segmentation (%) | W segmentation (%) |
|---|---|---|---|
| 0 | 0.0040 | **2.11** | **1.67** |
| 2 | 0.0037 | **1.97** | **1.81** |
| 3 | 0.0032 | **1.74** | 1.20 |
| 4 | 0.0019 | 1.41 | **0.90** |
| 5 | 0.0039 | **1.66** | 1.60 |
| 6 | 0.0027 | **1.99** | **1.88** |
| 7 | 0.0274 | **5.86** | 5.65 |
| 8 | 0.0040 | **2.28** | **1.67** |
| 9 | 0.0027 | **1.73** | **1.78** |
| 10 | 0.0041 | **2.51** | **1.99** |
| Mean | 0.0058 | **2.33** | **2.01** |

TABLE VIII
ABLATION EXPERIMENT OF PROPOSED MIXED FILTER

| Seq | Rotation error (deg/m) | BRISK % | AKAZE % | Super point % | SIFT % |
|---|---|---|---|---|---|
| 0 | 0.0040 | 1.84 | 1.92 | **1.92** | **1.67** |
| 2 | 0.0037 | 1.79 | 1.81 | **1.72** | **1.81** |
| 3 | 0.0032 | 1.72 | 1.40 | **2.00** | **1.20** |
| 4 | 0.0019 | 1.72 | 2.13 | 1.74 | 0.90 |
| 5 | 0.0039 | 1.92 | 1.59 | **1.91** | **1.60** |
| 6 | 0.0027 | 1.96 | 1.82 | **1.81** | **1.88** |
| 7 | 0.0274 | 5.60 | 5.77 | **5.03** | **5.65** |
| 8 | 0.0040 | 1.82 | 1.96 | **1.84** | **1.67** |
| 9 | 0.0027 | 1.59 | 1.59 | **1.53** | **1.78** |
| 10 | 0.0041 | 2.00 | 2.43 | **2.18** | **1.99** |
| Mean | 0.0058 | 2.2 | 2.24 | **2.17** | **2.01** |

To assess the performance of semantic segmentation, we conduct tests on sequences 00 and 02-10, as shown in Table VII. Taking sequence 00 as an example, the segmentation technique enhances both motion estimation and scale calculation results by utilizing dynamic object masks and incorporating rough road feature selection through road segmentation masks. This improvement is evident in the noticeable decrease in the RPE value.

Finally, we evaluate our method using four different Gaussian filter parameters. The sigma value plays a crucial role in regulating the smoothness of the scale queue utilized for computing the scale of the current frame. The outcomes, as displayed in Table VIII, demonstrate that a higher sigma value leads to smoother performance of MonoVO. Notably, when sigma equals 5, the average RPE decrease reaches 2.02%, indicating that the presence of noise caused by mismatching or imprecise triangulation is mitigated. The choice of sigma value depends on the desired trade-off between smoothness and responsiveness, considering the stability of scale changes within a shorter time frame.

## V. Conclusion

In this paper, we present an innovative hybrid methodology meticulously crafted to adeptly tackle the complex challenge of scale recovery within the realm of MVO. Our approach departs from convention by utilizing a computationally efficient segmentation model, specifically SegNeXt, for both ego motion estimation and ground point selection. On the other hand, our approach amalgamates a vote-based selection method and a road model-based selection method which significantly enhances the accuracy of road surface point selection. With the carefully selected road points, we estimate the road model using RANSAC and remove the noise using a mixed filter. Experimental results show that our method achieves accurate scale which significantly increase the performance of existing VO methods, including ORB-SLAM3 and MonoVO. By integrated with ORB-SLAM3 (monocular without loop closure), our method outperforms recent scale recovery method and six state-of-the-art VO systems. However, unadulterated MVO system faces pronounced vulnerability to factors like lighting fluctuations and rapid motion. A requisite degree of displacement and discernible features is essential during the initialization phase. In response to these challenges, our forthcoming research endeavors will prioritize the refinement of feature detection and matching methods, along with advancements in visual signal preprocessing for heightened accuracy.

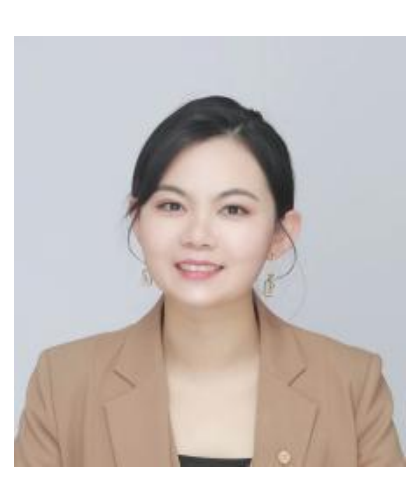

**Hui Zhang** (Member, IEEE) is currently affiliated with Shanghai Normal University. She received her Ph.D. degree from the Robotics and Artificial Intelligence Laboratory at Tongji University,

Shanghai, China, in 2022. From September 2017 to March 2018, she visited RAM-LAB, Robotics Institute, Hong Kong University of Science and Technology, Hong Kong. In 2015, she obtained her B.S. degree in automation from Anhui University, Hefei, China. Her research mainly focuses on visual localization for robotics with a focus on monocular visual odometry in the autonomous driving application. Additionally, she actively contributes to the academic community by serving as a reviewer for various computer science conferences, including ICRA and IROS.

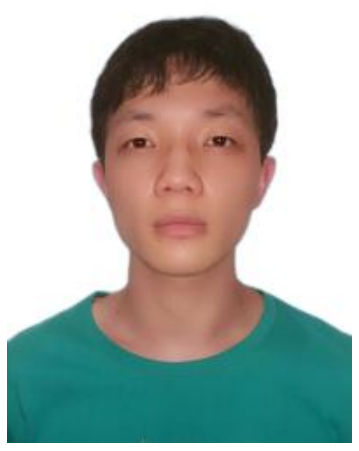

**Zhiyang Wu** (Member, IEEE) received his B.S. degree from Jiangsu University, Zhenjiang, China, in 2021. Between 2022 and 2023, he served as the primary contributor to the completion of the corporate delegation while pursuing his M.S. degree at Shanghai Normal University, Shanghai, China. His research primarily focuses on monocular visual odometry and visual SLAM. Additionally, in 2022, he achieved the Second Prize in the 17th National Smart Car Competition for College Students.

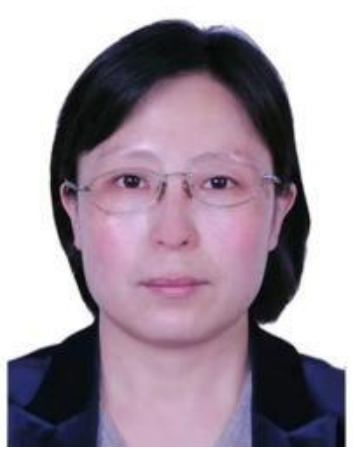

**Qianqian Shangguan** received the Ph.D. degree (2007) in Mechanical Design and Theory from Shanghai Jiao Tong University, B.E. (1996) and M.E. (1999) degrees in Mechanical and Electronic Engineering from Yanshan University. She serves as an associate professor in Shanghai Normal University. Her research interest mainly focuses on intelligent robots.

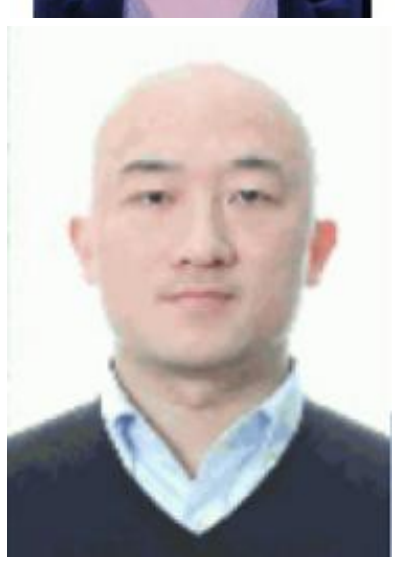

**Kang An** (Member, IEEE) received his Ph.D. degree in control theory and control engineering from Tongji University in 2014. He pursued joint Ph.D. training at Cornell University from 2010 to 2011, where he significantly contributed to the development of a bipedal robot that achieved a Guinness World Record in motion optimization. He is currently affiliated with Shanghai Normal University, his research focuses on intelligent robot control and perception, with particular interests in visual perception and intelligent robots.